%% file: coling2018.tex
\documentclass[11pt]{article}
\usepackage{coling2018}
\usepackage{times}
\usepackage{url}
\usepackage{latexsym}

\usepackage{multirow}
\usepackage{graphicx}
\usepackage{multicol}
\usepackage{algorithm}
\usepackage[noend]{algpseudocode}
\usepackage{amsmath}
\usepackage[font={footnotesize}]{caption}
\usepackage{enumitem}
\usepackage{amssymb}
\usepackage{subcaption}
\usepackage{comment}
\usepackage{color}

\newcommand\triviaqa{\textsc{TriviaQA}}
\newcommand\triviaqanop{\textsc{TriviaQA-NoP}}

\newcommand\nl[1]{\emph{``#1''}}
\newcommand\rightact{\textsc{Right}}
\newcommand\leftact{\textsc{Left}}
\newcommand\downact{\textsc{Down}}
\newcommand\uplact{\textsc{UpL}}
\newcommand\upract{\textsc{UpR}}
\newcommand\answeract{\textsc{Answer}}
\newcommand\stopact{\textsc{Stop}}
\newcommand\docqn{\textsc{DocQN}}
\newcommand\docqncoup{\textsc{DocQN-Coupled}}
\newcommand\dqncoup{\textsc{DQN-Coupled}}
\newcommand\sA{\mathcal{A}}
\newcommand\sS{\mathcal{S}}
\newcommand\sD{\mathcal{D}}
\newcommand\randomwalk{\textsc{RandomWalk}}
\newcommand\randompar{\textsc{RandomPara}}
\newcommand\tfidf{\textsc{Tf-Idf}}
\newcommand\doctfidf{\textsc{Doc-Tf-Idf}}
\newcommand\ensemble{Ensemble}
\newcommand\readtop{\textsc{ReadTop}}

\title{Learning to Search in Long Documents Using Document Structure}

\author{Mor Geva \\
  Tel Aviv University \\
  {\tt morgeva@mail.tau.ac.il} \\\And
  Jonathan Berant \\
  Tel Aviv University \\
  {\tt joberant@cs.tau.ac.il} \\}

\date{}

\begin{document}
\maketitle

\input{00_abstract}
\input{01_intro}
\input{02_setup}
\input{03_data}

\input{04_method}

\input{05_experimental_setup}
\input{06_results}

\input{07_related}
\input{08_conclusions}

\section*{Acknowledgments}
We thank Eunsol Choi from Washington University for helpful discussions and comments on the paper. This research was supported by the Yandex Initiative in Machine Learning and by The Israel Science Foundation grant 942/16. This work was completed in partial fulfillment for the Ph.D degree of the first author.

\bibliographystyle{acl}
\bibliography{all}

\newpage
\appendix
\input{09_supplemental}

\end{document}

%% file: 00_abstract.tex
\begin{abstract}
Reading comprehension models are based on recurrent neural networks that sequentially process the document tokens. As interest turns to answering more complex questions over longer documents, sequential reading of large portions of text becomes a substantial bottleneck. Inspired by how humans use document structure, we propose a novel framework for reading comprehension. We represent documents as trees, 
and model an agent that learns to interleave quick navigation through the document tree with more expensive answer extraction.
To encourage exploration of the document tree, we propose a new algorithm, based on Deep Q-Network (DQN), which strategically samples tree nodes at training time. 
Empirically we find our algorithm  improves question answering performance compared to DQN and a strong information-retrieval (IR) baseline, and that ensembling our model with the IR baseline results in further gains in performance.

\end{abstract}

%% file: 01_intro.tex
\section{Introduction}

\blfootnote{
    \hspace{-0.65cm}  
    This work is licensed under a Creative Commons 
    Attribution 4.0 International License.
    License details:
    \url{http://creativecommons.org/licenses/by/4.0/}
}

Reading comprehension (RC), the task of reading documents and answering questions about their content, has attracted immense attention recently. While early work focused on simple questions and short paragraphs \cite{hermann2015read,rajpurkar2016squad,trischler2017newsqa,onishi2016wdw}, current work is shifting towards more complex questions that require reasoning over long documents \cite{joshi2017triviaqa,hewlett2016wikireading,welbl2017constructing,kovcisky2017narrativeqa}.

Long documents pose a challenge for current RC models, as they are dominated by recurrent neural networks (RNNs) \cite{chen2016thorough,kadlec2016text,xiong2017dynamic}. RNNs process documents token-by-token, and thus using them for long documents is prohibitive. A common solution is to retrieve part of the document with an IR approach \cite{chen2017reading,clark2017simple} or a cheap model \cite{watanabe2017question}, and run an RNN over the retrieved excerpts. However, as documents become longer and questions become complex, two problems emerge: (a) retrieving all the necessary text with a one-shot IR method when performing complex reasoning becomes harder, and thus thousands of tokens are retrieved \cite{clark2017simple}. (b) Running even a cheap model over the document in its entirety becomes expensive \cite{choi2017coarse}.

Humans, in lieu of a mental inverted index, use document structure to guide their search for answers. E.g., the answer to \emph{``What high school did Leonard Cohen go to?''} is likely to appear in \emph{``Early life''}, while the answer to \emph{``How hot is it in Melbourne in July?''} is likely to appear in \emph{``Climate''}. In this work we investigate whether we can train a model to navigate through a document using its structure and find the answer, while reading only a small portion of the entire document.

\begin{figure}
\centering
\includegraphics[scale=0.33]{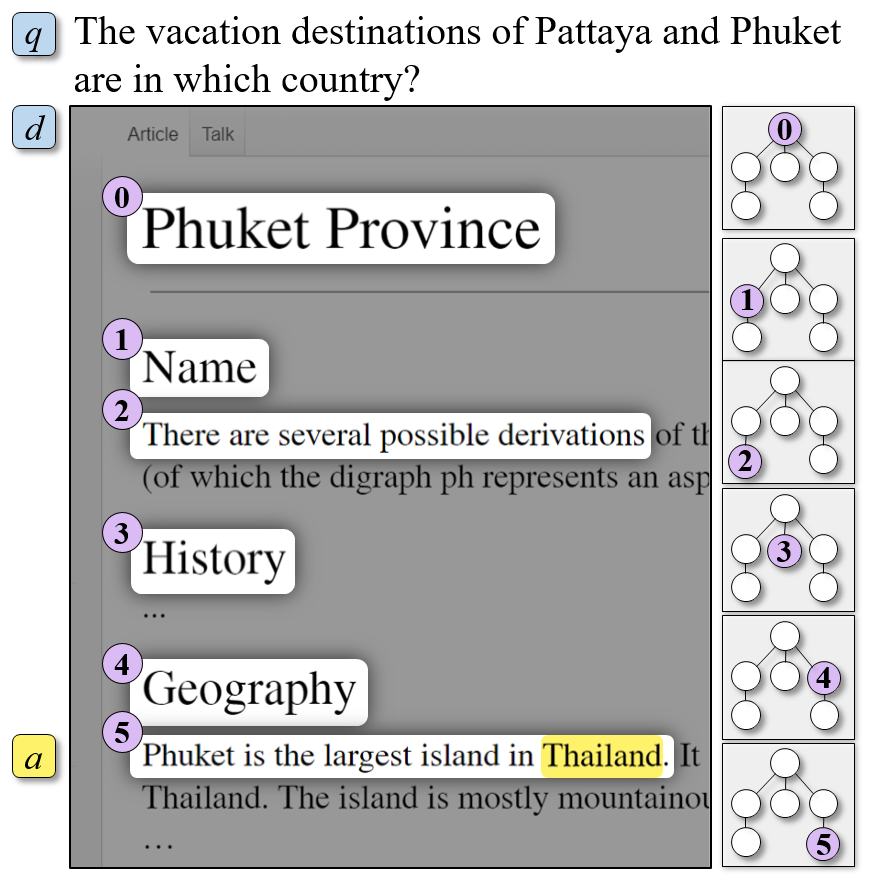}
\caption{An overview of our framework: an agent answers the question $q$ for a document $d$ by starting at the title, performing navigation actions until reaching the relevant paragraph, extracting the answer $a$, and then stopping.}
\label{fig:overview}
\end{figure}

We represent documents as trees and train an agent that navigates through the document tree until returning a final answer. Figure~\ref{fig:overview} illustrates this process. Our agent reads the question \nl{The vacation destinations of Pattaya and Phuket are in which country?} and starts navigation at the title of the document. After reading a paragraph, and skipping \nl{History}, it drills down to \nl{Geography} until finally halting at a paragraph that specifies the answer (\nl{Thailand}).
The agent observes at each step only a glimpse of the local text to determine its next action, which can be movement to a tree node, answering the question with a more expensive RC model, or terminating navigation. Thus, the agent consumes only a small fraction of the entire document.

Our training data is question-document-answer triplets, without gold navigation paths, and thus we train our model with the Deep Q-Network (DQN) algorithm. Because the dataset is biased towards answers appearing at the beginning of the document, the algorithm 
tends to stop early and does not explore the document well. To overcome this challenge, we
propose \docqn{}: a variant of DQN for tree navigation that improves exploration by sampling nodes from multiple parts of the tree. 


We evaluate the ability of our agent to navigate to paragraphs containing the answer on a variant of \triviaqa{} \cite{joshi2017triviaqa} and find that: 
(a) \docqn{} navigates better than DQN in documents both quantitatively and qualitatively.
(b)
While \docqn{} observes only 6\% of the document tokens, it outperforms an IR method in end-to-end QA performance. (c) An ensemble of \docqn{} and IR substantially improves both navigation and end-to-end QA performance over the ensemble components. 

To summarize, in this paper we ask: can an agent use document structure and learn to find answers for complex questions in long documents?
We propose a new model and training algorithm that overcomes an inherent bias in the data, answering the aforementioned question in the affirmative.
Our code and dataset are available at \url{https://github.com/mega002/DocQN}.

%% file: 02_setup.tex
\section{Problem Overview}
\label{sec:setup}
We work in the traditional RC setup, where we are given question-document-answer triplets $\{(q_i, d_i, a_i)\}_{i=1}^N$ as a training set, and aim to learn a function that finds the answer for an unseen question-document pair.
Unlike prior work, we assume documents are trees, where every tree node $u$ corresponds to a structural element and is labeled with text $l(u)$.
Specifically, the root is labeled with the document title, sections and subsections are labeled by their title, and paragraphs and sentences are labeled by the text they contain. In addition, we order all non-sentence tree nodes by a pre-order traversal (which corresponds to the linear order of text in the document), and denote the index of a node $u$ by $n(u)$. For sentence nodes, $n(u)$ is the index of their parent (a paragraph). 

Figure~\ref{fig:overview} shows an example tree, where for each node we show the relevant structural element and index (sentence nodes are not shown in the figure).

With this document representation, answering questions can be viewed as a Markov Decision Process (MDP),
where in each state the agent is located in a particular tree node, actions allow movement through the document tree, answering the question with a RC model, or stopping, and a reward is based on whether the agent locates a node that contains the answer.

Our agent interleaves actions that \emph{navigate} in the document, with an action that runs an RC model in a certain document position and \emph{extracts an answer}. Thus, the agent can decide to continue navigation after extracting a certain answer.
This is strictly more expressive than existing  approaches that combine IR with RC, where some text is retrieved exactly once before applying an RC model. 
As RC shifts to reasoning over complex questions, navigating and reading multiple parts of the document will become necessary. We show in Section~\ref{sec:experiments} this approach indeed improves QA performance.


%% file: 03_data.tex
\section{Data}
To test our framework, we capitalize on the recently-released \triviaqa{} dataset, which contains question-answer pairs, along with a small set of documents that (in almost all cases) contain the answer.
\triviaqa{} is suitable for our purposes as it is a large scale dataset, where questions are relatively complex and documents are fairly long. The dataset includes only raw text, and thus for every evidence document, we built a tree representation by extracting the html metadata from the corresponding Wikipedia page, and constructing the document structure from it.

\begin{figure}[t]
{\centering
\begin{minipage}[c]{0.48\textwidth}
   \centering
{\scriptsize
\begin{tabular}{llcc}
\hline
                          &           & \multicolumn{1}{l}{\triviaqa} & \multicolumn{1}{l}{\triviaqanop} \\ \hline
\multirow{2}{*}{Train}    & Questions & 61,888                        & 57,220                           \\ 
                          & Documents & 110,647                       & 99,315                          \\ \hline
\multirow{2}{*}{Development}      & Questions & 7,993                         & 7,336                            \\ 
                          & Documents & 14,229                        & 12,706                           \\ \hline
\multirow{2}{*}{Test}     & Questions & 7,701                         & 6,507                            \\ 
                          & Documents & 13,661                        & 10,481                           \\ \hline
\end{tabular}}
\captionof{table}{Data statistics for \triviaqa{} vs. \triviaqanop{}.}
\label{tab:data_statistics_a}
{\scriptsize
\begin{tabular}{l|c}
\multicolumn{2}{c}{}  \\
\hline
Average number of tokens  & 5590.7    \\ 
Average number of tree nodes  & 332.2    \\
Average number of high-level sections  & 6.6    \\ \hline
\end{tabular}}
\captionof{table}{Data statistics for the evidence documents of \triviaqanop{}.}
\label{tab:data_statistics_b}
\end{minipage}
\quad
\begin{minipage}[c]{0.46\textwidth}
    \includegraphics[scale=0.37]{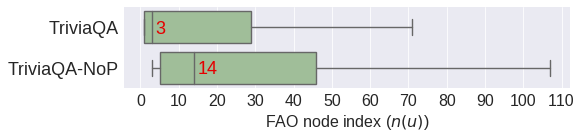}
\captionof{figure}{FAO node index distribution of \triviaqa{} and \triviaqanop{} (median values in red).}
\label{fig:fao_distribution_a}
\centering 
{\scriptsize
\begin{tabular}{lcc}
\multicolumn{3}{c}{}  \\
\hline
  & \triviaqa & \triviaqanop \\ \hline
Questions & 86.7\% & 83.8\% \\ \hline
Question-document pairs & 69.7\% & 66.9\% \\ \hline
\end{tabular}}
\captionof{table}{Portion of answerable samples in a random subset of the training set, which contains 278 questions and 475 question-document pairs.}
\label{tab:manual_analysis}
\end{minipage}}
\end{figure}

Because our goal is to investigate whether a model can learn to search through a document, it is important that a non-negligible fraction of the questions require navigation through the document. However, in Wikipedia each document starts with a preface that summarizes the document, and thus often contains the answer. Consequently, a model that ignores the question and document and always stops in the first paragraph is likely to obtain good performance. 
Figure~\ref{fig:fao_distribution_a} shows
the distribution of the first answer occurrence (FAO) in a document in \triviaqa{} over the node indices $n(u)$ (x-axis), where all tree nodes, except sentences are considered (see Section~\ref{sec:setup}). 
We find that in most question-document pairs the FAO is in the first few paragraphs, and that in 60\% of the cases it is in the preface section.


To alleviate this heavy bias, we derive a new dataset, termed \triviaqanop{}, where we remove the preface section from every document. After removing the preface, 2,144 out of 77,582 (2.8\%) questions and 6,124 out of 138,537 (4.4\%) question-document pairs are left without an answer and are removed. To further reduce the number of cases where an answer can not be inferred from a document, we drop question-document pairs where: (a) the answer appears only in titles; (b) the answer is a single-character; (c) the FAO node index is $>700$ (in most cases the answer is an item in a list). Finally, the dataset includes 91.6\% of the questions and 88.4\% of the question-document pairs from the original dataset. We provide full statistics on the dataset in Tables~\ref{tab:data_statistics_a} and \ref{tab:data_statistics_b}.

To verify that questions remain answerable in \triviaqanop{} after removing the preface, we perform manual analysis on a random sample of 278 questions and 475 documents from the training set (Table~\ref{tab:manual_analysis}).
We find that the portion of answerable questions and question-document pairs remains high and is reduced by less than 3\% in comparison to \triviaqa{} .
This demonstrates that indeed, for most questions and documents, the context necessary for answering the question also appears in the document body and not only in the preface. 

Figure~\ref{fig:fao_distribution_a} shows the FAO node index distribution in \triviaqanop{}. We observe, compared to \triviaqa{}, that the first occurrence of an answer is much more spread out across the document, and that the median increases from 3 to 14, which will require more navigation from the agent.
However, even in \triviaqanop{} answers tend to appear at the beginning of the document, because document content is usually organized by importance. This bias results in an exploration challenge for our training algorithm, which we will address in Section~\ref{sec:method}.

%% file: 04_method.tex
\section{Method} \label{sec:method}

In this section, we describe a model for the agent and a training algorithm based on DQN \cite{mnih2015human}. Specifically, we introduce a tree-sampling strategy, which addresses the exploration challenge stemming from the bias towards answers early in the document.

\subsection{Framework} \label{subsec:framework}

We represent the MDP as a tuple $(\sS, \sA, R, T)$, where $\sS$ is the state space, $\sA$ is the action space, $R(s,a)$ is a reward function, and $T(s,a)$ is a deterministic transition function. Our model implements an action-value function $Q(s,a)$, which takes a state $s \in \sS$ and returns a value for every action $a \in \sA$. This function defines a policy $\pi(s) = \arg\max_{a \in \mathcal{A}} Q(s,a)$. We now describe the state space, actions and reward function.

\begin{table}
\centering
    \scriptsize
    \begin{tabular}{p{4cm}|c}
    & \bf Example \\ \hline 
    $o$ & \nl{Phuket Province Name} \\[0.3ex] \hline
    $\phi_n^1$: \texttt{height} &  2 \\[0.3ex] \hline
    $\phi_n^2$: \texttt{depth} &  1 \\[0.3ex] \hline
    $\phi_n^3$: \texttt{h\_dist\_start} & 0 \\[0.3ex] \hline
    $\phi_n^4$: \texttt{h\_dist\_end} & 2 \\[0.3ex] \hline
    $\phi_n^5$: \texttt{parent.h\_dist\_start} & 0 \\[0.3ex] \hline
    $\phi_n^6$: \texttt{parent.h\_dist\_end} & 0 \\[0.3ex] \hline
    $\phi_n^7$: \texttt{navigation step} & 1 \\ 
    \end{tabular}
   \caption{An example observation $o$ and list of navigation features $\phi_n$ for node 1 in Figure \ref{fig:overview}. The features \texttt{height} and \texttt{depth} correspond to distance from the farthest leaf and root, respectively (\texttt{height} is 2 since sentence nodes are omitted from the figure). \texttt{h\_dist\_start} and \texttt{h\_dist\_end} measure the horizontal distance from the first and last child of the node's parent. \texttt{navigation step} is a counter for the number of performed steps.}
   \label{tab:observation_components}
\end{table}

\paragraph{States} Given a tree node $u$, a state is a tuple $s^u=(q,o,z,\phi_n,\phi_z)$, where $q$ is the question, $o$ is an observation, $z$ is an answer prediction, and $\phi_n$, $\phi_z$ are navigation and answer prediction features. 
An observation $o = (o_1, \dots, o_{|o|})$ is a sequence of tokens produced by recursively concatenating the first $k$ tokens of text in the label $l(u)$ to the observation of $u$'s parent. 
An answer prediction $z=(z_1, \dots, z_{|z|})$ is the sequence of tokens that were extracted by an RC model, if an RC model was already run on $l(u)$ (and a null token otherwise).
The answer prediction features $\phi_z = (z_e,\, z_l,\, z_n)$ provide information on the distribution over answer spans provided by the RC model, which reflects its confidence: $z_e$ is the entropy of the distribution, $z_l$ is the logit value for $z$, and $z_n$ is the number of tokens in $l(u)$.
Navigation features $\phi_n$ provide information on the relative location of $u$ in the document.
An example observation and full list of navigation features are given in Table \ref{tab:observation_components}. 

Note that the state $s$ does not depend on the history of visited tree nodes.\footnote{except for \texttt{navigation step}, which can be approximated by the shortest path from the root to any node.} While incorporating history could be beneficial, a memory-less model enables us to explore tree-sampling strategies, which is important for training (Section \ref{subsec:docqn}).

\begin{figure}[ht]
\centering
    \includegraphics[scale=0.08]{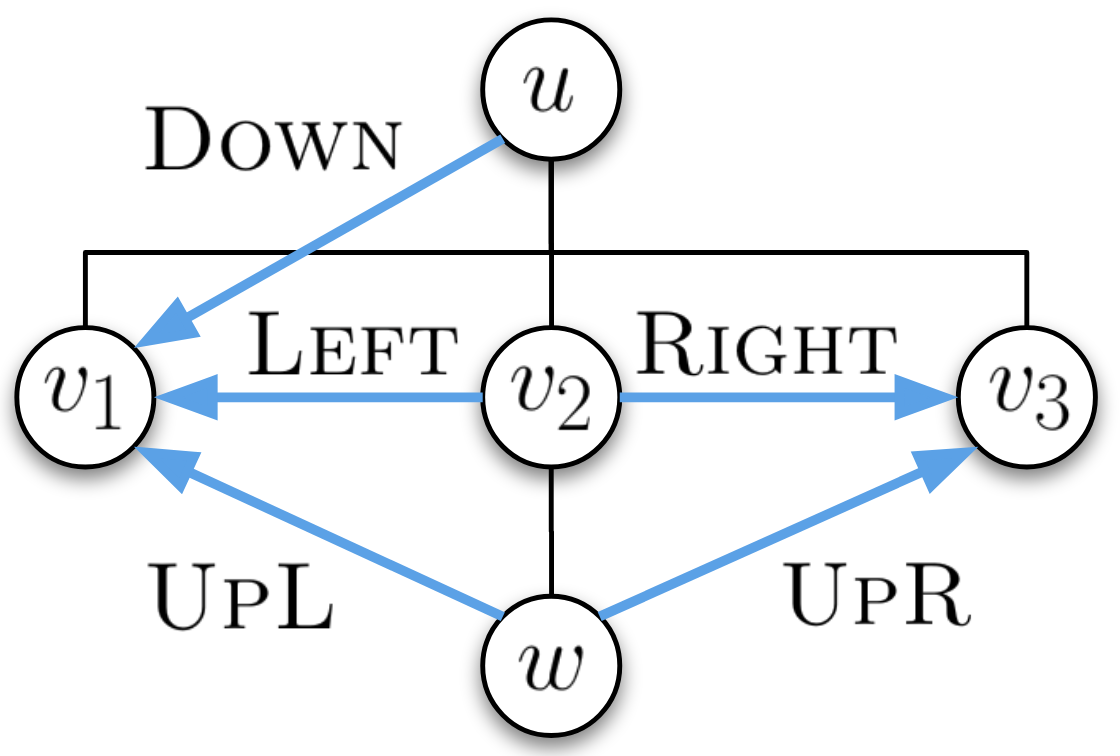}
    \caption{Movement actions in our environment.}
    \label{fig:movement-actions}
\end{figure}

\paragraph{Actions} We define the following set of actions $\mathcal{A}$. Let $u$ be a node with an ordered list of children $(v_1, v_2, v_3)$, and $w$ be a child of $v_2$. We define five movement actions (Figure~\ref{fig:movement-actions}), where \downact{} moves from $u$ to its first child $v_1$, \rightact{} moves from $v_2$ to $v_3$, and \leftact{} moves from $v_2$ to $v_1$. Because moving upwards reaches a node we already visited, we define \upract{}, which moves from $w$ to $v_3$, and \uplact{}, which moves from $w$ to $v_1$. If an illegal action is chosen (e.g., \leftact{} from $v_1)$, then the agent stays in its current position. 
The action \answeract{} returns an answer (and a distribution over spans) by running a RC model on $l(u)$, unless $u$ is a sentence, in which case it is run on the paragraph containing $u$.
After \answeract{}, the agent can resume navigation. The action \stopact{} also returns the answer given the current node $u$, but also terminates navigation.


\paragraph{Reward}
Our goal is to develop an agent that can navigate in the document, and thus 
we define the reward based on whether the agent stops in a node that contains the gold answer text (this is noisy, because the answer might be there sporadically).
While a simple reward would be an indicator for whether the agent stopped at a correct node, such a reward would not capture the proximity of the agent to the answer. Moreover, we would like  to consider the overall document length, rewarding successful navigations in long documents.
Therefore, we define the following reward:

\begin{equation*}
\small
r(s,a) = \begin{cases}
2 & a=\stopact, |n(u) - n(u^*)| = 0 \\
1 - \frac{|n(u) - n(u^*|)}{\max_u n(u)} & a=\stopact, |n(u) - n(u^*)| > 0 \\
- 0.06 & a = \answeract \\
- 0.02 & a \not \in \{\answeract, \stopact\}
\end{cases}
\end{equation*}
where $u$ is the node where the agent is located and $u^*$ is the closest tree node that contains the answer ($n(u)$ is the node index as defined above). Thus, when stopping, the reward is proportional to the distance to the closest answer location given the document length. An additional reward is given if navigation is successful, and a penalty is given for any other action, to encourage shorter trajectories. We further penalize the \answeract{} action to discourage frequent usage of the RC model.

\subsection{\docqn{}: DQN with Tree Sampling}
\label{subsec:docqn}
Training the navigation model is based on DQN \cite{mnih2015human}. In DQN, at every step an agent at state $s_t$ selects an action $a_t$ using $\epsilon$-greedy policy, given the current action-value function $Q_\theta(s_t,a_t)$ parameterized by $\theta$. The agent observes a reward $r_t$ and a state $s_{t+1} = T(s_t, a_t)$ and adds a transition $(s_t, a_t, r_t, s_{t+1})$ to a replay memory buffer $\mathcal{D}$ that holds a large number of recent transitions. The parameters are then optimized so that the action-value function  matches better the observed reward. This is done by sampling a batch of random transitions from $\mathcal{D}$ and minimizing the regression loss 
\begin{equation} \label{eq:dqn}
(r_t + \gamma \max_{a'} Q_{\hat{\theta}}(s_{t+1}, a') - Q_\theta(s_t, a_t))^2,
\end{equation}
where $\hat{\theta}$ are the parameters of a \emph{target network}, which is a periodic copy of $\theta$ that is not optimized, and $\gamma$ is a discount factor.

We also add some of the recent enhancements to DQN \cite{hessel2017rainbow}, which have proved to be useful in our setup. Specifically we implement \emph{Double Q-Learning} \cite{van2016deep}, \emph{Prioritized Experienced Replay} \cite{schaul2015prioritized}, and \emph{Dueling Networks} \cite{wang2016dueling}.


\paragraph{Reducing bias with \docqn}
The DQN algorithm contains episodes, where in each episode the agent is placed at an initial state $s_0$, from which it starts taking actions. In our setup, this state corresponds to the root of the document tree. Because the data has a bias towards answers appearing at the beginning of the document, the agent learns that stopping early improves the reward and is stuck at a local minimum, where it ceases exploration. 
Examining Figures~\ref{fig:fao_distribution_a} and~\ref{fig:fao_distribution_b}, we observe that DQN learns to stop very early in the document compared to the FAO node index distribution of \triviaqanop{}, amplifying the bias in the data.


To address this issue, we suggest \docqn{}, a variant of DQN aimed at increasing exploration when navigating in a document structure. \docqn{} capitalized on two properties. First, DQN is an off-policy algorithm that trains from transitions $(s_t, a_t, r_t, s_{t+1})$. Second, our model is memory-less, and thus we can sample any node $u$ and compute the corresponding state $s^u$. Therefore, we modify DQN, and instead of initializing every episode with $s_0$ and performing a sequence of actions, we sample states from distributions that explore the document better. By exploring transitions from across the document, the model learns from more distant parts of the document.

\begin{figure}[t]
\centering
\includegraphics[scale=0.22]{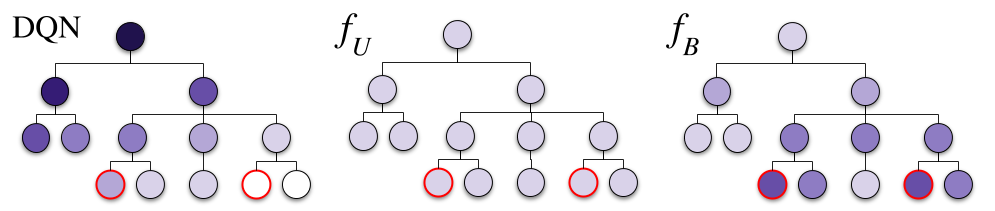}
\caption{Illustration of the different state distributions for a specific documents tree. Nodes with darker color have higher sampling probability, and nodes marked in red are paragraph nodes containing the answer.}
\label{fig:state_distributions}
\end{figure}
\begin{figure}[t]
{\begin{minipage}{0.44\textwidth}
    \captionof{algorithm}{\textbf{\docqn{}}}
    \label{alg:docqn}
    {\footnotesize
    \begin{algorithmic}[1]
    \State Let $f$ be a distribution over tree nodes    
    \State Initialize replay memory $\sD$ and parameters $\theta,\hat{\theta}$
    \For{ episode $=1,M$ }
    \If{ random() $< \varepsilon_s $ } \label{line:sample_far}
    \For{ $i=1 \dots K$ }
    \State sample node $u \sim f$ and generate $s^u$
    \State sample action $a$ from $s^u$ ($\epsilon$-greedily).
    \State $r \leftarrow R(s^u,a), s' \leftarrow T(s^u,a)$
    \State store $(s^u, a, r, s')$ in $\sD$
    \EndFor
    \Else \label{line:sample_usual}
    \State Initialize start state $s_0$
    \For{ $t=0 \dots T-1$ }
    \State sample action $a$ from $s_t$ ($\epsilon$-greedily).
	\State $r \leftarrow R(s_t,a), s_{t+1} \leftarrow T(s_t,a)$    
    \State store $(s_t, a, r, s_{t+1})$ in $\sD$
    \EndFor
    \EndIf
    \State $\textsc{UpdateParams}(\theta, \hat{\theta}, \mathcal{D})$ (Equation \ref{eq:dqn})   
    \EndFor
    \end{algorithmic}
    }
\end{minipage}
\begin{minipage}{0.48\textwidth}
\includegraphics[scale=0.235]{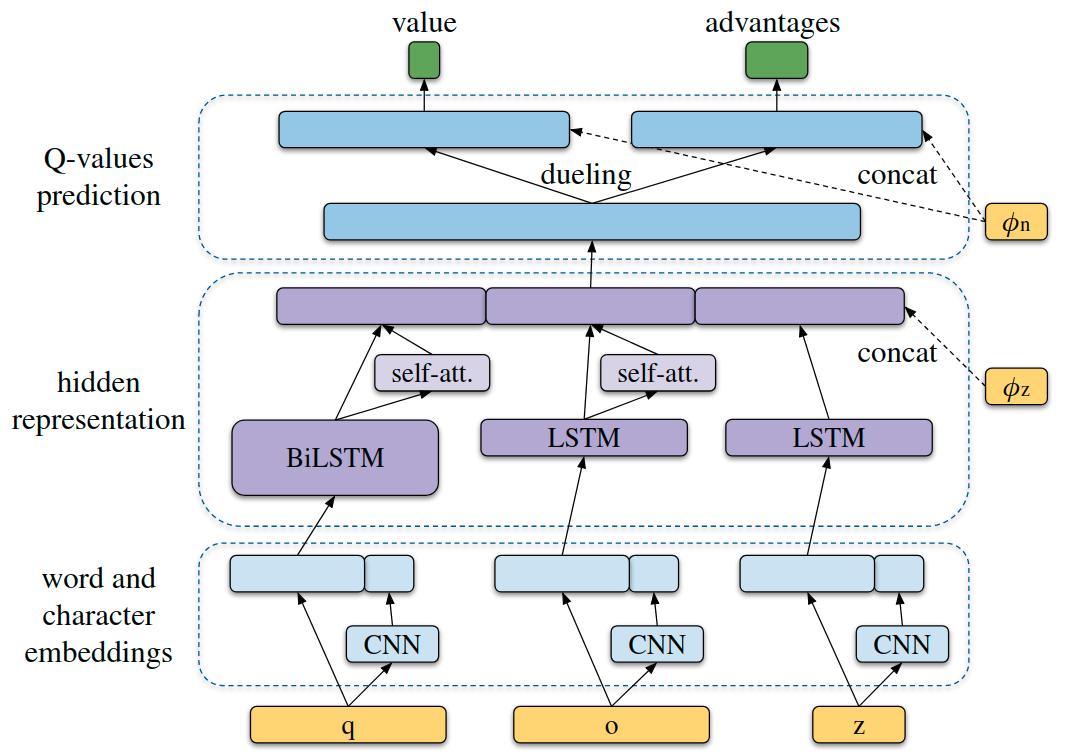}
\captionof{figure}{High-level overview of the network architecture.}
\label{fig:network_diagram}
\end{minipage}}
\end{figure}


Algorithm~\ref{alg:docqn} provides the details of \docqn{}. The algorithm initializes the replay memory buffer, the model and target network parameters, and chooses a distribution $f$ over tree nodes. At the beginning of every episode, with probability $\epsilon_s$ (line~\ref{line:sample_far}),\footnote{$\epsilon_s$ is annealed from $1$ to $0.5$ during training.} we sample $K$ state transitions using $f$ and store them in $\sD$, and with probability $(1-\epsilon_s)$ (line~\ref{line:sample_usual}), we start at the initial state $s_0$ and sample a trajectory as in DQN. Parameters are updated as in DQN by sampling transitions from the replay buffer and minimizing Equation~\ref{eq:dqn}. If $f$ samples nodes from various locations in the document, the algorithm will explore better and will not getting stuck at its beginning. We consider the following instantiations of $f$:
\begin{enumerate}[topsep=0pt,itemsep=0ex,parsep=0ex]
\item $f_U$: \emph{Uniform sampling} over nodes, except we discourage sampling sentences by uniformly sampling with probability $0.2$ a leaf (sentence), and probability $0.8$ an inner node.
\item $f_B$: \emph{Backward sampling}: We uniformly sample a paragraph node $p$ that contains the gold answer, then uniformly a number of actions $B\in \{1,2,3\}$, and  perform $B$ random movement actions from $p$ to output a node. This results in a node that is ``close'' to the answer $a$, and can be viewed as similar to bi-directional search or backward search \cite{lao2015learning}.
\end{enumerate}

Figure~\ref{fig:state_distributions} shows the distribution over tree nodes for a specific document tree where the answer appears in two paragraph nodes (ignoring sentence nodes). We see that sequential sampling (as in DQN) puts most of the probability mass close to the root, uniform sampling is uniform (across paragraphs, as sentence nodes are omitted), and backward sampling is concentrated close to answer nodes. 
This illustrates how different distributions result in different exploration strategies.

\paragraph{Network Architecture} \label{subsec:neural-model}

We briefly describe the neural architecture (Figure~\ref{fig:network_diagram}) and provide full details in the supplementary material.
As explained in Section~\ref{subsec:framework}, the input to the network is the state $s$, which comprises the question tokens $q$, observation tokens $o$, answer prediction tokens $z$ and features $\phi_n$, $\phi_z$.
The question, observation and answer prediction tokens are encoded with pre-trained word embeddings and trained character embeddings, where character embeddings are followed by a convolutional layer with max pooling, yielding a single vector per token. Each token is then represented by concatenating the word embedding with the character embedding.

Question tokens, observation tokens, and answer to are then fed into a BiLSTM and LSTM \cite{hochreiter1997lstm} respectively and the LSTM outputs are compressed to a single vector through self attention \cite{cheng2016long}, resulting in one vector for $q$ and one for $o$.
Answer prediction tokens are fed into an LSTM, where the last hidden state is concatenated to the features $\phi_z$, thus creating a third vector for the answer prediction.

We concatenate these three vectors, and pass them through a one layer feed-forward network that then branches to two networks according to the Dueling DQN architecture \cite{wang2016dueling}. In each branch we also concatenate the navigation features $\phi_n$.
One branch predicts the value of the state $V_\theta(s) \in \mathbb{R}$, and the other branch predicts the advantage of every action $A_\theta(s,a) \in \mathbb{R}$ for every possible action $a$. The output of the network is $Q_\theta(s,a) = V_\theta(s) + (A_\theta(s,a) - \frac{1}{\sA}\sum_{a'}A_\theta(s,a'))$ as in \newcite{wang2016dueling}.

%% file: 05_experimental_setup.tex
\section{Experimental Evaluation}
\label{sec:experiments}

Our experimental evaluation aims to answer the following questions: (a) Can document structure be used to learn to navigate to an answer in a document? (b) How does \docqn{} compare to DQN? (c) How does \docqn{} compare to IR methods that observe the entire document?

\subsection{Experimental Setup}

We evaluate on \triviaqanop{}. Because our focus is on the navigation ability of the agent, we train a single RC model and fix it in all experiments.
Specifically, we download \textsc{RaSoR}   \cite{lee2017rasor,salant2018contextualized},\footnote{\url{https://github.com/shimisalant/RaSoR}} and exactly follow the procedure described by the authors of \triviaqa{} for training a RC model, i.e., we train \textsc{RaSoR} on the first $400$ tokens of each document in \triviaqanop{}.
As a sanity check for our RC model, we also train and evaluate \textsc{RaSor} on the original \triviaqa{} dataset. Indeed, \textsc{Rasor} obtains $48.6\%$ EM and $53.4\%$ F1, which is substantially higher than the baseline reported by \newcite{joshi2017triviaqa}.


\paragraph{Evaluation}
We use two metrics: First, we measure \emph{navigation accuracy}, i.e., for a question-document pair, whether 
a method returns text containing a gold answer (if the agent stops at a sentence node we evaluate the encompassing paragraph). Because questions in \triviaqanop{} often have more than one evidence document, we also measure \emph{aggregated navigation accuracy}, where we give credit if the agent navigated correctly in any of the documents. This gives  performance assuming an oracle that always chooses the best document for the question. Because the test set in \triviaqa{} is hidden, we evaluated navigation accuracy on the development set only.

In addition, we measure end-to-end QA performance with the official Exact Match (EM) and F1 metrics. The EM metric measures the percentage of predictions that match exactly any of the answer aliases, and the F1 metric measures the average overlap between the prediction and answer. To aggregate evidence from multiple documents we follow \newcite{joshi2017triviaqa} and define the score of an answer to be the sum of probabilities for that answer across all documents.

\paragraph{Models}
We compare the following models: 
\begin{itemize}[topsep=0pt,itemsep=0ex,parsep=0ex,leftmargin=\parindent]
\item \textbf{\docqn{}}: This is our main model, where we use the sampling distribution $f_{U+B} = 
\frac{1}{2} f_U + \frac{1}{2} f_B$.
\item \textbf{DQN}: DQN algorithm without state sampling.
\item \textbf{\textsc{\{DocQN$\mid$DQN\}-Coupled}}: A less expressive version of the model, where the actions \textsc{ANSWER} and \textsc{STOP} are coupled, i.e., the agent can use the RC model exactly once and then stops. This is similar to a setup where retrieval is performed once and not interleaved with navigation.
\item \textbf{\randomwalk} An agent that selects an action at each step uniformly at random.
\item \textbf{\randompar} An agent that randomly selects a non-sentence tree node.  
\item \textbf{\doctfidf{}}:
An IR baseline, where a paragraph is selected based on its tf-idf score. We implement the tf-idf scheme of \textsc{DocQA} \cite{clark2017simple}, which is a high-performing system on \triviaqa{}. Here, idf is computed for each paragraph in the context of the current document, and the paragraph with highest cosine similarity to the question is chosen.
Note that \doctfidf{} processes the entire document (up to tens of thousands of tokens), while \docqn{} processes only a small fraction.
\item \textbf{\tfidf{}}:
Vanilla tf-idf, where the idf score is computed from all documents in \triviaqanop.
\item \textbf{\ensemble}: We ensemble \docqn{$f_{U+B}$} with \doctfidf{} in two ways: (1) for finding the final answer, we aggregate scores as described above, that is, we sum the probabilities from both models over all documents and choose the answer with highest score. (2) For navigation, we simply tune on the development set a threshold $l$, where we take the prediction of \docqn{$f_{U+B}$} if it stopped at a node with index $\leq l$ and use \doctfidf{} otherwise.
\item \textbf{\readtop{}}: 
Following \newcite{joshi2017triviaqa}, a model that runs the RC model on the first $800$ tokens of the document. Note that 
running the RC model on the text retrieved by \docqn{} involves consuming far fewer tokens, namely, \textsc{RaSoR} consumes only $160$ tokens on average.
\end{itemize}
We report the value of all hyper-parameters in the supplementary material (all fixed without tuning).

%% file: 06_results.tex
\subsection{Results}

\begin{minipage}[t]{.525\textwidth}
\vspace{0pt}
\includegraphics[scale=0.375]{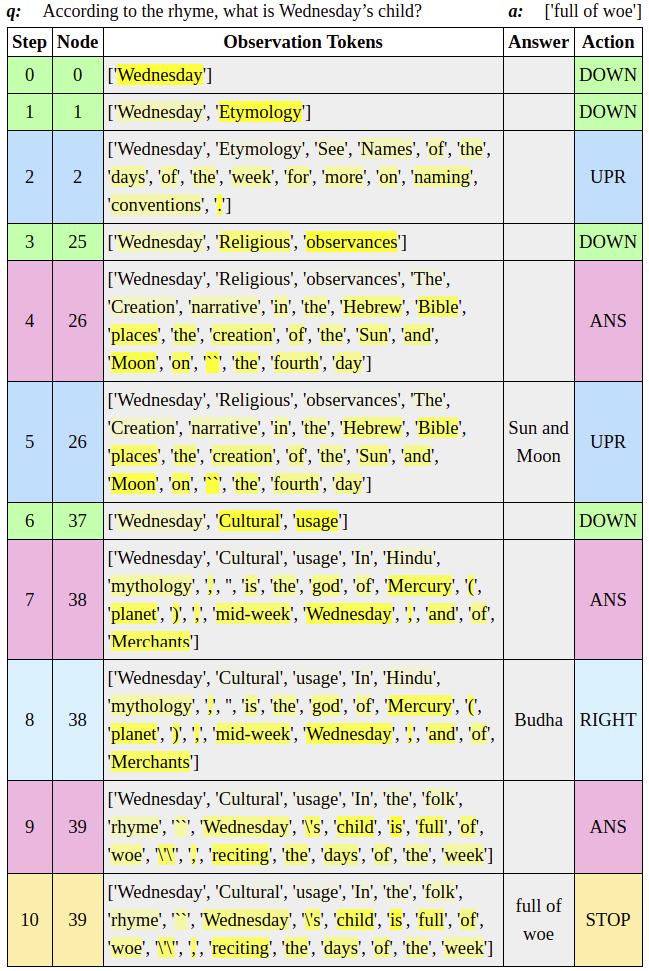}
\captionof{figure}{Navigation example of \docqn{}. }
\label{fig:nav_exm1}
\vspace{10pt}
\end{minipage}
\quad
\begin{minipage}[t]{.465\textwidth}
\vspace{0pt}
\begin{minipage}[t]{\dimexpr \textwidth-2\fboxsep-2\fboxsep\relax}%
\includegraphics[scale=0.35]{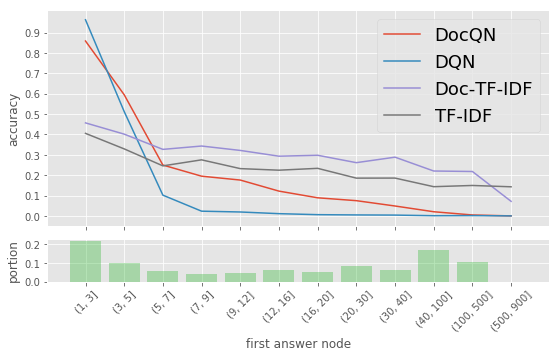}
\captionof{figure}{Model performance (top) and portion of samples in the development set (bottom) as functions of the node index of the FAO.}
\label{fig:accuracy_per_fao}
\end{minipage}
\begin{minipage}[c][\dimexpr 0.25\textheight-20\fboxsep-2\fboxrule\relax][t]{\dimexpr \textwidth-2\fboxsep-2\fboxsep\relax}%
\vspace{30pt}
\includegraphics[scale=0.38]{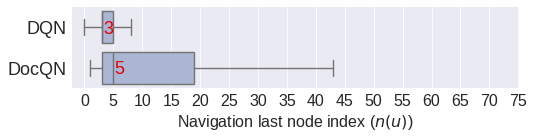}
\captionof{figure}{Distribution of node index at navigation stopping point (median value in red) for DQN and \docqn{}.}
\label{fig:fao_distribution_b}
    \end{minipage}
\end{minipage}

Tables ~\ref{tab:navigation_results} and ~\ref{tab:answer_results} show the results of our experiments.
Focusing on navigation accuracy, we see that randomly walking or choosing a paragraph yields low performance. Vanilla \tfidf{} performs considerably better than the random baselines, but is outperformed by all other models. Comparing DQN to \docqn{}, we see that \docqn{} outperforms DQN. Allowing DQN and \docqn{} to run the RC model during navigation improves their performance, but the accuracy gain is substantially higher for \docqn{} (2.3\% accuracy and 3.1\% aggregated accuracy) than for DQN (0.4\% accuracy and 0.7\% aggregated accuracy).
\doctfidf{}, which has access to the entire document outperforms \docqn{}, which consumes on average 6\% of the entire document. Nonetheless, the two models obtain the same aggregated accuracy.
This good performance of \doctfidf{} shows that in \triviaqanop{} answer paragraphs share a lot of lexical material with the question.
Importantly, 
an ensemble of \doctfidf{} with \docqn{} substantially improves the overall performance, reaching accuracy of 35\% and aggregated accuracy of 48\%. This is in sharp contrast to an ensemble with DQN, where for any value of $l$, \doctfidf{} performs better on its own.

\begin{table}[t]
    \begin{minipage}[t]{.45\linewidth}
      \centering
{\scriptsize
\begin{tabular}{|lcc|}
\multicolumn{3}{c}{}  \\
\hline
  & Dev. & Dev. Agg. \\ \hline
\multirow{1}{1.8cm}{\randomwalk} & 1.8 & 3.1 \\ \hline
\multirow{1}{1.8cm}{\randompar} & 13.4 & 20.9 \\ \hline
\multirow{1}{3.6cm}{\dqncoup{}} & 26.8 & 38.8 \\ \hline
\multirow{1}{3.6cm}{DQN} & 27.2 & 39.5 \\ \hline
\multirow{1}{3.6cm}{\docqncoup{} } & 28.1 & 40.5 \\  \hline
\multirow{1}{3.6cm}{\docqn{}} & 30.4 & 43.6 \\  \hline
\multirow{1}{3.6cm}{\tfidf} & 25.3 & 35.4 \\ \hline
\multirow{1}{1.8cm}{\doctfidf} & 32.4 & 43.5 \\ \hline
\multirow{1}{3.6cm}{Ensemble (threshold, $l=5$)}  & \textbf{35.0} & \textbf{48.0} \\ \hline
\end{tabular}
}
\caption{Navigation accuracy for all models on the development set of \triviaqanop{}.
}
\label{tab:navigation_results}
    \end{minipage}%
    \qquad
    \begin{minipage}[t]{.45\linewidth}
      \centering
{\scriptsize
\begin{tabular}{|lcccc|}
\hline
 & \multicolumn{2}{c}{Development} & \multicolumn{2}{c|}{Test} \\
 & EM & F1 & EM & F1 \\ \hline
\multirow{1}{1.8cm}{DQN} & 21.7 & 26.5 & 19.1 & 24.1 \\ \hline
\multirow{1}{1.8cm}{\docqn{}} & 23.6 & 27.9 & 21.0 & 25.5 \\ \hline
\multirow{1}{1.8cm}{\doctfidf} & 21.4 & 27.1 & 18.2 & 23.5 \\ \hline
\multirow{1}{3.6cm}{Ensemble (threshold, $l=5$)}  & 26.8 & 32.0 & 24.2 & 29.4 \\ \hline
\multirow{1}{2.8cm}{Ensemble (answer)}  & \textbf{28.4} & \textbf{33.4} & \textbf{25.4} & \textbf{30.5} \\ \hline \hline
\multirow{1}{1.8cm}{\textsc{ReadTop}} & \textbf{32.5} & \textbf{36.7} & \textbf{28.1} & \textbf{32.4} \\ \hline
\end{tabular}
}
\captionsetup{width=7.5cm}
\caption{End-to-end QA performance of all models on the development and test sets of \triviaqanop{}. For the ensemble model, we choose the best threshold of $l=5$.}
\label{tab:answer_results}
    \end{minipage} 
\end{table}

Examining end-to-end performance, \docqn{} again outperforms DQN, but \docqn{} is now better than \doctfidf{}. This suggests that when \doctfidf{} selects a paragraph with the answer, it is often difficult to extract with the RC model. Again, the ensembles leads to a dramatic increase in performance, showing that \doctfidf{} and \docqn{} are complementary.
However, \readtop{}, which consumes the first $800$ tokens of each document compared to $\sim160$ for \docqn{}, substantially outperforms all navigation models. This shows that the bias for answers at the beginning of documents is still strong in \triviaqanop{}.

To further elucidate the differences between navigation models, Figure~\ref{fig:accuracy_per_fao} shows navigation accuracy of different models and the proportion of samples for different node indices of the FAO. We see that DQN outperforms \docqn{} when the answer is at the top of the document, but \docqn{} dominates DQN when the answer is further down, showing that \docqn{} learns to find answers deeper in the document. \doctfidf{} has a more balanced navigation accuracy across the document, which explains why an ensemble of \doctfidf{} with \docqn{} works, as they are complementary to one another.

\paragraph{Analysis}

Figure~\ref{fig:nav_exm1} shows a navigation example, which includes the navigation step, node index, observation $o$, and the action $a$ taken. For the observation we also highlight the attention distribution from the self-attention component.
In this figure the question is about culture, and we see the agent going into multiple sections, reading them, running the RC model and continuing forward, until finally stopping at a paragraph that contains the answer. We provide many more  examples in the supplementary material.

Table~\ref{tab:navigation_properties} highlights some differences between DQN and \docqn{}. 
\docqn{} has longer trajectories, and stops at sentence nodes more frequently, suggesting it reads the document at a finer granularity. Additionally, it leverages the RC model to collect more information and confidence during navigation, by choosing the action ANSWER more frequently.
Both models consume less than 7\% of the entire document.
Figure~\ref{fig:fao_distribution_b} illustrates the navigation stopping point, and shows that \docqn{}  navigates deeper into the document.

\begin{table}
\centering
\scriptsize
\begin{tabular}{|p{2cm}|p{2.4cm}|p{2.4cm}|}
\multicolumn{3}{c}{}  \\
\multicolumn{3}{c}{}  \\
\hline
 & DQN & \docqn{} \\ \hline
Path length & avg. 7.7, range [1,36] & avg. 15.2, range [3,100] \\ \hline
Answer predictions & avg. 2.8 & avg. 4.1 \\ \hline
Tokens consumed & 3.4\% & 6.2\% \\ \hline
Stopping node  & 0.5\% title, 2.2\% headline, 89.0\% paragraph, 8.3\% sentence & 0\% title,  0.4\% headline, 75.1\% paragraph, 24.5\% sentence \\ \hline
\end{tabular}
\captionof{table}{Comparing navigation properties of DQN and \docqn{} on the development set. The average and range of navigation path length in steps, relative amount of consumed tokens, and distribution of stopping node types.}
\label{tab:navigation_properties}
\end{table}

%% file: 07_related.tex
\section{Related Work}

Handling the challenges of reasoning over multiple long documents is gaining fast momentum recently \cite{shen2017reasonet}. As mentioned, some approaches use IR for reducing the amount of processed text \cite{chen2017reading,clark2017simple}, while others use cheap or parallelizable models to handle long documents \cite{hewlett2017accurate,swayamdipta2018multi,wang2018r3}. Searching for answers while using a trained RC model as a black-box was also implemented recently in \newcite{wang2018evidence}, for open-domain questions and multiple short evidence texts from the Web.
Another thrust has focused on skimming text in a sequential manner \cite{yu2017skim}, or designing recurrent architectures that can consume text quickly \cite{bradbury2017quasi,seo2018neural,campos2018skip,yu2018fast}. However, to the best of our knowledge no work has previously applied these methods to long documents such as Wikipedia pages.

In this work we use \triviaqanop{} for evaluation of our navigation based approach and comparison to an IR baseline. While there are various aspects to consider in such evaluation setup, our choice of data was derived mainly by the requirements for long and structured context. Recently, several new datasets such as \textsc{WikiHop} and \textsc{NarrativeQA} were published. These datasets try to focus on the tendency of RC models to match local context patterns, and are designed for multi-step reasoning. \cite{welbl2017constructing,wadhwa2018towards,kovcisky2017narrativeqa}.

Our work is also related to several papers which model an agent that navigates in an environment to find objects in an image \cite{ba2015multiple}, relations in a knowledge-base \cite{das2018go}, or documents on the web \cite{nogueira2016end}.



%% file: 08_conclusions.tex
\section{Conclusions}
We investigate whether document structure can be leveraged to train an agent that finds answers to questions in long documents while reading only a small fraction of it. We show that an agent that reads 6\% of the document can improve QA performance compared to an IR method that utilizes the entire document, and that ensembling the two substantially improves performance.
We also present \docqn{}, an algorithm that promotes better exploration of the document, and show it outperforms DQN qualitatively and quantitatively.

Our approach represents a conceptual departure from previous methods for reading long documents, as it interleaves searching for an answer in the document with extracting the answer from a particular paragraph, which we show improves both navigation and QA performance.
We expect that as RC models tackle longer documents that require reasoning and reading text that is spread in multiple parts of the document,
models that can efficiently navigate and collect evidence will become more and more crucial. Our agent provides a first step in this important research direction.

%% file: 09_supplemental.tex
\appendix

\section{Supplementary Material}
\label{sec:supplemental_a}

\subsection{Network Architecture Details}
Here, we elaborate on the network architecture, which was briefly described in the paper. Given an input state $s=(q,o,z,\phi_n,\phi_z)$, we denote by $q=(q_1,...,q_n)$, $o=(o_1,...,o_m)$ and $z=(z_1,...,z_r)$ the question tokens, observation tokens and answer prediction tokens, respectively.

\paragraph{Word-level embedding}
For every $i=1,...,n$, every $j=1,...,m$ and every $k=1,...,r$, we create an embedding of the input tokens $q_i, o_j, z_k$ in two steps. First, we embed every token with a pre-trained GloVe word embedding matrix $W_w$: 
\begin{align*}
e_{q_i}^w = W_w \cdot q_i \quad;\quad e_{o_j}^w = W_w \cdot o_j \quad;\quad e_{z_k}^w = W_w \cdot z_k
\end{align*}

Next, we apply character-level embeddings with a learned embedding matrix $W_c$. The embedded characters are then summarized with a max-pooling convolutional neural network:
\begin{align*}
&e_{q_i}^c = \mathrm{CNN}(\{W_c \cdot q_{ix}\}_{x=1}^{|q_i|}) \\
&e_{o_j}^c = \mathrm{CNN}(\{W_c \cdot o_{jx}\}_{x=1}^{|o_j|}) \\
&e_{z_k}^c = \mathrm{CNN}(\{W_c \cdot o_{kx}\}_{x=1}^{|z_k|})
\end{align*}

Concatenation of the two components yields the final word-level embeddings:
\begin{align*}
e_{q_i} = [e_{q_i}^w \,;\, e_{q_i}^c] \quad;\quad e_{o_j} = [e_{o_j}^w \,;\, e_{o_j}^c] \quad;\quad e_{z_k} = [e_{z_k}^w \,;\, e_{z_k}^c]
\end{align*}

\paragraph{Question sequence encoding}
The question is encoded with a BiLSTM, where for every timestamp $i$, the forward and backward outputs are concatenated:
\begin{gather*}
\{u_{q_1}, ..., u_{q_n}\} = \mathrm{BiLSTM}(e_{q_1}, ..., e_{q_n})\\
u_{q_i} := [\overrightarrow{\mathrm{LSTM}}(\overrightarrow{h}_{q_{i-1}}, e_{q_i}) \;;\; \overleftarrow{\mathrm{LSTM}}(\overleftarrow{h}_{q_{i+1}}, e_{q_i})]
\end{gather*}
The outputs are then summarized with a self-attention:
\begin{equation*}
h_q = \sum_{i=1}^n{\alpha_i u_{q_i}}
\end{equation*}
where the coefficients $\alpha_1,...,\alpha_n$ are obtained by feeding the outputs to a two-layer feed-forward network, and normalizing the output logits with softmax:
\begin{align*}
a_i = \mathrm{FFNN}(u_{q_i}) \quad;\quad \alpha_i = \frac{e^{a_i}}{\sum_{k=1}^n{e^{a_k}}}
\end{align*}

\paragraph{Observation sequence encoding}
The observation sequence encoding $h_o$ is obtained in an analogous manner to $h_q$, except that we use a $\mathrm{LSTM}$ rather than a $\mathrm{BiLSTM}$.

\paragraph{Answer prediction encoding}
The answer prediction encoding $h_z$ is obtained by running the answer prediction tokens through a LSTM, and concatenating the last hidden state with the input feature vector $\phi_z$:
\begin{gather*}
\{u_{z_1}, ..., u_{z_r}\} = \mathrm{LSTM}(e_{z_1}, ..., e_{z_r})\\
u_{z_k} := \overrightarrow{\mathrm{LSTM}}(\overrightarrow{h}_{z_{k-1}}, e_{z_k})\\
h_z = [u_{z_r}\;;\;\phi_z]
\end{gather*}

\paragraph{State representation}
The final state representation is formed by concatenating the encoded question $h_q$ with the encoded observation $h_o$. Concretely:
\begin{equation*}
h_s = [h_q\;;\; h_o\;;\; h_z]
\end{equation*}
The input node features $\phi_o$ are concatenated to an upper layer, as we describe next , to increase their weights in the final predictions. We have found that incorporation of these features in this way accelerates the navigation learning process.
 
\paragraph{Q-values prediction}
We implement a Dueling DQN architecture, where the final Q-values are composed of a state value $V(s)$ prediction and advantage predictions $A(s,a)$ for every possible action $a$. Denoting by FFNN a single-layer feed-forward neural network, predictions are derived as follows:
\begin{gather*}
v_0 = \mathrm{FFNN}(h_s) \\
v_1^V = \mathrm{FFNN}(v_0) \;;\; v_1^A = \mathrm{FFNN}(v_0) \\
v_2^V = \mathrm{FFNN}([v_1^V \;;\; \phi_n]) \;;\; v_2^A = \mathrm{FFNN}([v_1^A \;;\; \phi_n])
\end{gather*}
Where $v_2^V \in \mathbb{R}$, $v_2^A \in \mathbb{R}^{|\sA|}$, and $\phi$ are the navigation features.
The final Q-values prediction is obtained by averaging:
\begin{equation*}
v^Q = v_2^V + \Big(v_2^A - \frac{\sum_{i=1}^{|\sA|} v_{2i}^A}{|\sA|}\Big)
\end{equation*}

\subsection{Hyper Parameters}
Table \ref{tab:parameters} summarizes the hyper-parameters used for building and training the \docqn{} models. 

\begin{table}[t]
\centering
{\scriptsize
\begin{tabular}{llc}
\hline
   &  \multicolumn{1}{l}{Parameter}  & \multicolumn{1}{c}{Value} \\ \hline
\multirow{2}{1cm}{Network}
                          & Maximum node observation length & 20 tokens \\
                          & Maximum observation length & 120 tokens \\
                          & Word embedding dimension & 300 \\ 
                          & Character embedding dimension & 20 \\ 
                          & Convolution filter size & 5 \\ 
                          & BiLSTM and LSTM hidden dimension & 300 \\
                          & First feed-forward layer dimension & 512 \\
                          & Second feed-forward layer dimension & 256 \\
                          & Dropout rate & 0.2 \\  
\hline
\multirow{3}{1cm}{Training}    & RMSprop learning rate & 0.0001 \\ 
    & Batch size & 64 \\
    & Target network period & 10K steps \\ 
                          & Initial memory size & 50K transitions \\
                          & Maximal memory size & 300K transitions \\
                          & Action sampling $\epsilon$ & $1.0\rightarrow 0.1 $ \\
                          & Discount factor $\gamma$ & $0.996 $ \\
                          & Prioritization usage $\alpha $ & 0.6 \\
                          & Prioritization importance sampling $\beta$ & $0.4 \rightarrow 1.0 $ \\
\hline
\multirow{2}{1cm}{Sampling}   & State sampling $\epsilon_s$ & $1.0\rightarrow 0.5 $ \\
   & Annealing steps for $\epsilon_s$ & $1.2$M steps \\
   & Sampling repetitions $K$ & 5 \\ 
   & Maximum navigation length $T$ (train) & 30 steps \\ 
   & Maximum navigation length $T$ (test) & 100 steps \\ 
   & Interpolation coefficient for $f_{U+B}$ & 0.5 \\ 
\hline
\end{tabular}
}
\caption{\docqn{} hyper-parameters}
\label{tab:parameters}
\end{table}

\subsection{Navigation Examples}

\begin{figure}
\centering
\begin{minipage}[t]{0.45\textwidth}
\includegraphics[scale=0.32]{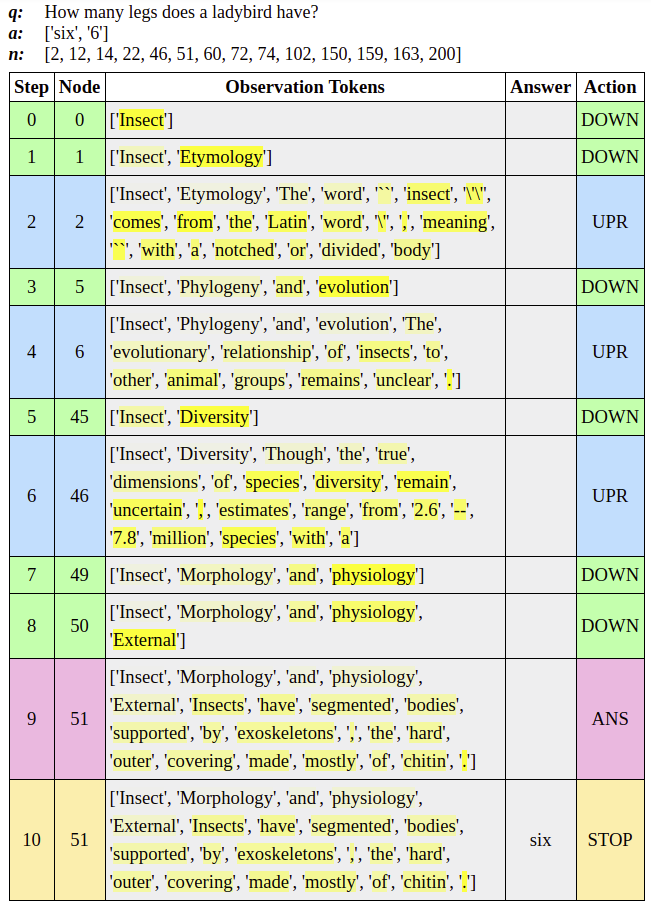}
\captionof{figure}{Navigation example 1. Although the document is more general than the question subject and the answer appears multiple times across the document, the agent finds the correct context for answering the question.}
\label{fig:nav_exm2}
\end{minipage}
\qquad
\begin{minipage}[t]{0.45\textwidth}
\includegraphics[scale=0.30]{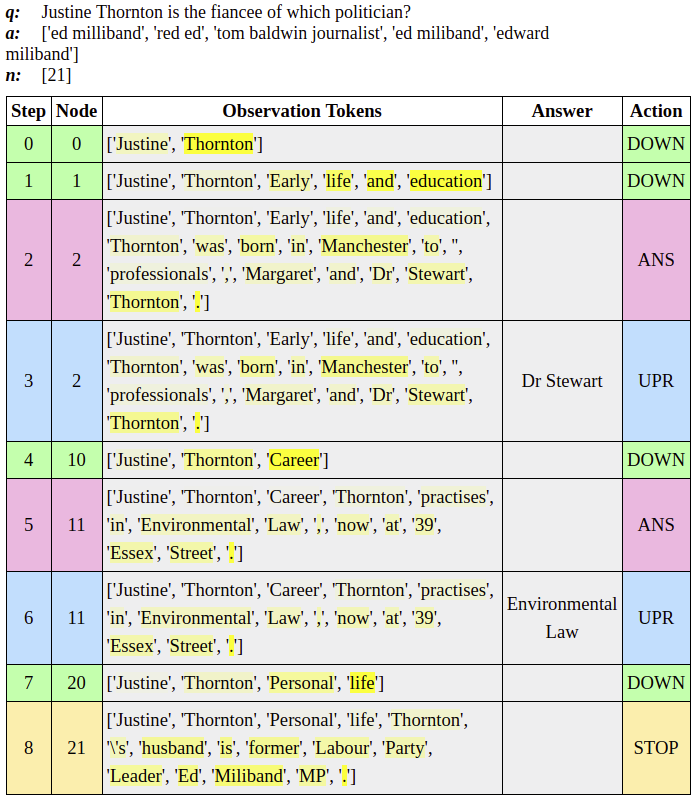}
\captionof{figure}{Navigation example 2. The agent explores several sections before reaching the most relevant one.}
\label{fig:nav_exm3}
\end{minipage}
\end{figure}

\begin{figure}
\centering
\begin{minipage}[t]{0.45\textwidth}
\includegraphics[scale=0.33]{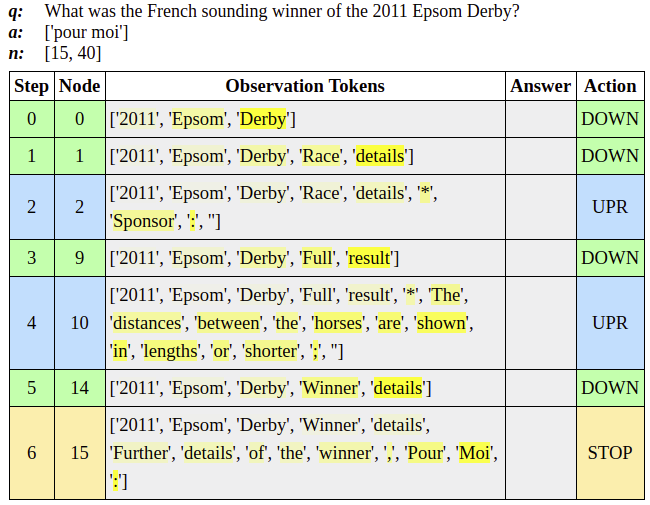}
\captionof{figure}{Navigation example 3. The agent quickly navigates to the answer, without running the RC model.}
\label{fig:nav_exm4}
\end{minipage}
\qquad
\begin{minipage}[t]{0.45\textwidth}
\includegraphics[scale=0.31]{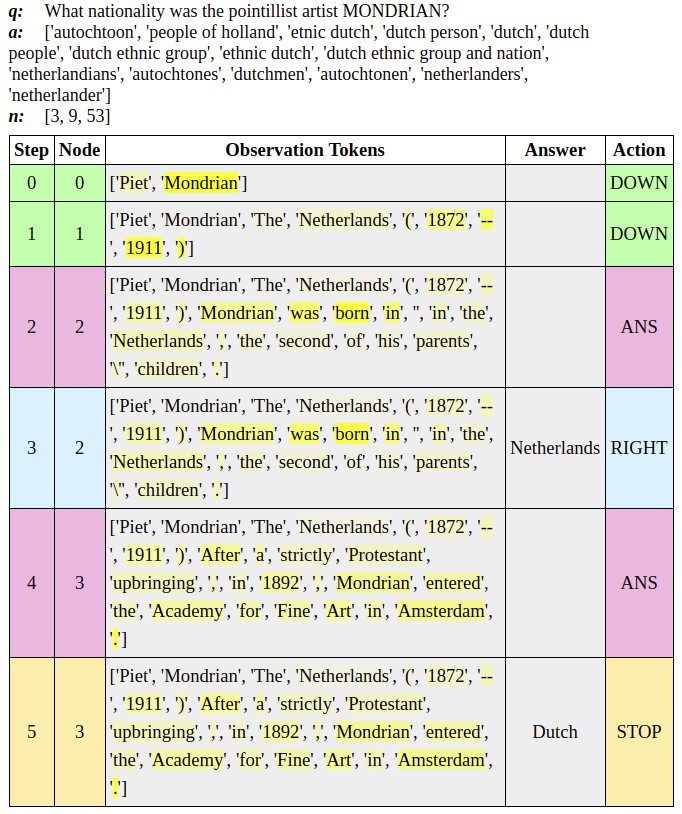}
\captionof{figure}{Navigation example 4. The agent leverages the RC model and decides to stop after observing enough information to answer.}
\label{fig:nav_exm5}
\end{minipage}
\end{figure}

\begin{figure}
\centering
\begin{minipage}[t]{0.51\textwidth}
\includegraphics[scale=0.36]{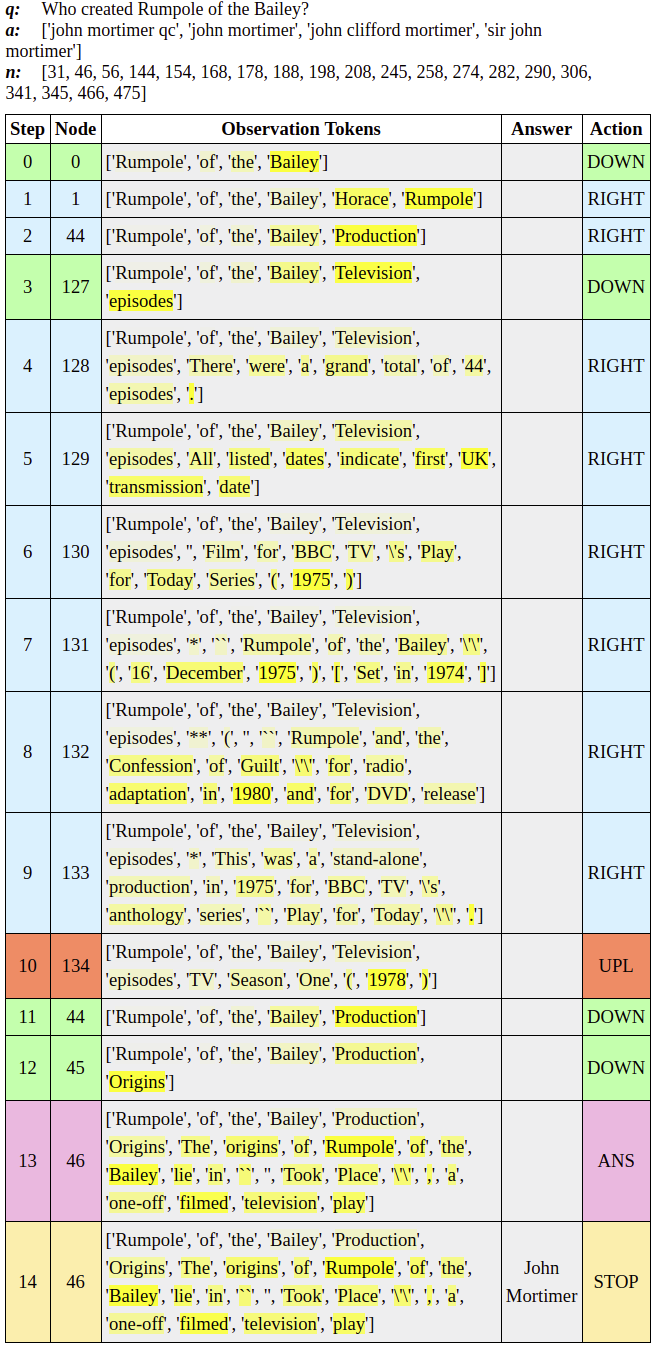}
\caption{Navigation example 5. After reading through the "Television episodes" section, the agent goes back to the "Production" section to find the answer.}
\label{fig:nav_exm6}
\end{minipage}
\end{figure}

Figures \ref{fig:nav_exm2},\ref{fig:nav_exm3},\ref{fig:nav_exm4},\ref{fig:nav_exm5},\ref{fig:nav_exm6} show sample navigations, performed by the \docqn{} model. In each example, $q,a,n$ denote the question, answer aliases and answer node numbers. The observation tokens are highlighted according to the self-attention weights, given by the model. By choosing the action ANS, the agent executes the RC model to obtain an answer prediction, which is part of the observation in the next step.

